\documentclass[pdflatex,sn-apa,oneside]{sn-jnl}

\usepackage{graphicx}%
\usepackage{multirow}%
\usepackage{amsmath,amssymb,amsfonts}%
\usepackage{amsthm}%
\usepackage{mathrsfs}%
\usepackage[title]{appendix}%
\usepackage{xcolor}%
\usepackage{textcomp}%
\usepackage{manyfoot}%
\usepackage{booktabs}%
\usepackage{algorithm}%
\usepackage{algorithmicx}%
\usepackage{algpseudocode}%
\usepackage{listings}%

\newcommand{\seq}[2]{#1\,{\tiny$\pm$#2}}

\theoremstyle{thmstyleone}%
\theoremstyle{thmstyletwo}%

\theoremstyle{thmstylethree}%

\begin{document}

\title[Article Title]{Measuring AI Accountability Through Argumentation Analysis: Can Model Reasoning Withstand Scrutiny?}


\author*{\fnm{Daan R.} \sur{Henselmans}}\email{daan@aithos.org}

\author{\fnm{Derck W.E.} \sur{Prinzhorn}}\email{derck@aithos.org}

\author*{\fnm{Arno} \sur{Libert}}\email{arno@aithos.org}

\affil{\orgname{Aithos Research Foundation}, \orgaddress{\street{Keizersgracht 62}, \city{Amsterdam}, \postcode{1015 CS}, \country{the Netherlands}}}

\abstract{AI oversight methods rely on ground truth for validation, but what constitutes appropriate AI behavior is contested. This leaves evaluation of moral reasoning in LLMs and debate-based oversight implicitly avoiding realistic ambiguity. We investigate an alternative standard designed to function despite such ambiguity: structural quality of the defence a model can mount for its verdicts in response to critical questions, measured through a four-phase dialectical protocol grounded in Walton's theory of argumentation schemes and Govier's criteria for argument cogency. The protocol is adaptive to different frames of reasoning, extends beyond multiple-choice framing, and treats both the reasoning that precedes a verdict and its post-hoc justification. Across nine frontier models and 200 high-ambiguity MoralChoice items---$6{,}778$ judge-scored cells, validated against $89.6\%$ inter-judge agreement on the binary failure judgment---models defend their reasoning well above the rubric minimum on every dimension. Failure mass concentrates on grounds and sufficiency, and correlates with epistemic hedging rather than argument length. Reasoning is \emph{better} defended than post-hoc justification, on every model and every Govier dimension. The scheme a model presents in its justification differs from the one it reasoned with on a substantial share of dilemmas  ($\geq 20\%$ per model), despite value-based practical reasoning dominating both tracks. The protocol catches strictly indefensible defences (self-contradiction, false premises), and it surfaces difficulties in characterizing the role of retraction in AI alignment, suggesting a need for more situated evaluations.}




\maketitle

\section{Introduction}

In real life, morally charged decisions cannot be checked against an answer key. Justifying a controversial choice requires taking responsibility, stating your reasons, and allowing others to legitimately challenge them \citep{habermas1990,forst2012}. As LLMs are deployed in roles affecting people's lives, like companionship, medical advice, or resource allocation, their actions too must do more than meet a behavioral target, and be justifiable to those they affect. But before we can evaluate whether an AI system can justify itself, we need to be able to distinguish a genuine moral argument from a plausible-sounding but incoherent rationalization.

\citet{haas2026} formalize the distinction between moral \emph{performance}---producing morally appropriate outputs---and moral \emph{competence}---producing them on the basis of morally relevant considerations. To distinguish genuine moral reasoning from a ``facsimile process'' that produces correct outputs without meaningful grounds, they advocate for adversarial evaluations to test whether a model's verdicts are produced by appropriate structural correspondence. This paper proposes an implementation using argumentation theory: rather than testing whether a model \emph{reaches} a verdict stated to be defensible, we test whether it \emph{can defend} whatever verdict it reaches when probed with the critical questions its own argument invites.

We adapt Walton, Reed and Macagno's theory of argumentation schemes \citep{walton2008}. When a model justifies a verdict, we classify the argumentation scheme it instantiates, generate that scheme's critical questions, and score the model's answers on Govier's criteria for argument cogency \citep{govier2010}. The methodology is adaptive to moral framing: a utilitarian justification and a deontological one can both score highly if each withstands scrutiny. What we measure is whether a justification meets structural requirements of cogent argumentation---a standard independent of normative targets, compatible with pluralistic alignment \citep{sorensen2024}.

A central design choice separates pre-verdict \emph{reasoning} from post-verdict \emph{justification} and evaluates both independently, motivated by evidence that chain-of-thought traces do not reliably reflect the causal processes behind model outputs \citep{lanham2023,turpin2023} and that reasoning steps can both reveal and conceal misaligned intentions \citep{chua2025}. Comparing the two tracks exposes whether the public justification reflects the underlying reasoning or is a mere rhetorical polish---a discrepancy that signals reduced accountability.

\section{Background and Related Work}

\subsection{Accountability and justification}

Justifiability as an ethical criterion for the legitimacy of institutions and the norms governing them is a recurring concept in modern political philosophy \citep{forst2012,habermas1996,rawls1993,scanlon1998}. Several proposals center it as an alignment target: \citet{gabriel2025} argue that AI systems must be regulated by principles based on fair processes amenable to public justification, while contractualist proposals follow Scanlon in the notion that systems should be governed by norms diverse stakeholders would endorse rather than by aggregated preferences \citep{zhixuan2025, levine2026}.

\citet{habermas1990} distinguishes \emph{monological} justification---conducted through solitary reasoning, addressed to nobody in particular---from justification whose argumentation addresses those affected, allowing contestation and coordination. \citet{levine2026} propose an alignment target which would answer to a ``soft'' version of this program: preemptively modeling affected parties' counterarguments. However, current LLM simulations still fail to meet deliberation and representation standards on ethically contested questions \citep{flechtner2026}, and it should be noted Habermas endorses the ``hard'' version embodied by actual discourse with affected parties, though the two are not mutually exclusive.

\subsection{Evaluating moral outputs vs.\ moral reasoning}
 
A growing body of work documents morally consequential failures in LLM output: agentic misalignment \citep{lynch2025}, sycophantic agreement over truthful responses \citep{sharma2023}, emergent deception \citep{chua2025}. The evaluation literature spans ethics benchmarks for normative moral judgments \citep{hendrycks2020,jin2022} and simulation environments that observe behavior under moral pressure \citep{chiu2025,hendrycks2021,pan2023}. These evaluations share a focus on moral \emph{performance}---producing morally appropriate outputs---rather than moral \emph{competence}---producing them on the basis of morally relevant considerations \citep{haas2026,kilov2025}. Moral competence is necessary to ground the public trust that legitimate deployment requires \citep{manzini2024}.

It can be difficult to distinguish quality moral reasoning from justificatory rhetoric. Mechanistic interpretability can identify causal links between internal representations in smaller models \citep{lindsey2025} but not sufficiently to confirm a verdict rests on morally salient considerations. A justification may be requested from the model itself, but frontier systems are now more persuasive than prepared, incentivized expert human debaters in live exchange \citep{hackenburg2026,salvi2025}, potentially making it difficult to identify invalid justifications. Chain-of-thought monitoring has been proposed as a distinct and fragile oversight opportunity \citep{korbak2025}, yet providers increasingly encrypt reasoning traces, and the summaries standing in for them can be unfaithful to the reasoning they summarize \citep{panfilov2026}. Even where traces are available, reasoning can rationalize injected biases without surfacing them \citep{turpin2023}, faithfulness shows inverse scaling with model capability \citep{lanham2023}, and reasoning models fine-tuned on narrow malicious behaviors emit both overt deceptive planning and benign-sounding rationalizations that evade monitoring \citep{chua2025}.

Debate between AI systems has been proposed as a proxy to identify truth in ambiguous circumstances \citep{irving2018}. However, its potential use in contested domains is undercut by its need for a ground truth to validate the answer \citep{irvingaskell2019}. The empirical case inherits this: debate helps identify the correct answer on tasks like reading comprehension, mathematics, and coding \citep{michael2023, kenton2024, khan2024}, but these setups require assigning one debater an incorrect position, which makes it less persuasive \citep{carro2025} and prone to concede \citep{khan2024}. Debate studies on ambiguous moral questions examine epistemic positions \citep{prasad2025}, prior beliefs \citep{carro2025}, and interaction protocols \citep{sachdeva2026}, but not the justifiability of verdicts.

\citet{haas2026} identify the ``facsimile problem'' as the fundamental obstacle: LLM architectures do not guarantee that the processes generating moral outputs are structurally analogous to the moral reasoning those outputs appear to reflect. They advocate adversarial, disconfirming evaluations that disentangle genuine reasoning from surface-level pattern matching. Our methodology implements this program through argumentation theory, with each scheme's critical questions serving as structured adversarial probes: a facsimile process should fail when the specific inferential steps of its argument are challenged.

\subsection{Argumentation theory and argument quality}

Walton, Reed and Macagno's theory of argumentation schemes provides the structural backbone of our methodology \citep{walton2008}. An argumentation scheme is a stereotypical pattern of human reasoning---argument from consequences, practical inference, argument from analogy---each associated with \emph{critical questions} that identify the conditions under which the argument's presumption can be defeated. Rather than inferring dialectical accuracy from the initial argument, our protocol poses critical questions matching the instantiated scheme and scores the response. Because the questions attach to the scheme, the objections owed are fixed to the argument the arguer chooses to make. Walton argues that ethical argumentation is a species of goal-directed practical reasoning, best evaluated through case-based dialectical exchange in which defeasible arguments are challenged and defended \citep{walton2003}---the basis for our four-phase protocol. Walton's framework has been applied computationally to classify schemes in text \citep{feng2011} and to model dialogue protocols \citep{gordon2006}, but to our knowledge has not been used as an evaluation methodology for LLM moral reasoning.

Govier's ARG conditions provide our primary evaluative criteria \citep{govier2010}. A cogent argument has premises that are \emph{acceptable} (reasonable to believe), \emph{relevant} (bearing on the conclusion), and that together provide adequate \emph{grounds} (sufficient reason to accept the conclusion). These derive from informal logic and target real-world argumentation rather than formal deductive validity \citep{johnson1977}. The conditions are partially ordered: relevance failure entails grounds failure---if premises do not bear on the conclusion they cannot supply adequate reason---but grounds can fail on its own when premises are acceptable and relevant yet insufficient, making grounds the most demanding and last to clear \citep{govier2010}. An argument can fail acceptability by overstating premises, relevance by trading on emotively loaded but non-bearing claims, or grounds by over-generalizing, independently of the normative position it argues for. Motivated by the distinction between local argument quality and global argumentation quality \citep{blair2012, wachsmuth2017, eemerengrootendorst2004}, we complement per-question ARG scoring with a global equivalent, which evaluates how the argumentation relates to itself as a whole (\S\ref{sec:methods-rubrics}).

Alongside argument quality, we track commitment stability as a diagnostic measure. Revising one's position in light of new considerations is a legitimate part of ethical argumentation \citep{walton2003}, but balance is key: both an arguer who keeps hedging and evading commitment and one who never retracts however strong the counterargument make dialogue pointless \citep{waltonkrabbe1995}. The first failure has been observed empirically in debate, where dishonest debaters evade being pinned down and steer away from the weak parts of their argument \citep{barneschristiano2020}.

\section{Methods}
\label{sec:methods}
 
We measure the argumentative moral competence of a language model by subjecting every response to a four-phase dialectical protocol (\S\ref{sec:methods-protocol}) grounded in Walton's argumentation schemes \citep{walton2008}. Rather than scoring the verdict a model reaches, the protocol scores the \emph{defence} it can mount when systematically probed with critical questions fixed to the scheme it instantiated. The output is a profile of Govier ARG scores \citep{govier2010} for the model's \emph{reasoning} and its \emph{justification}.
 
\subsection{Task and dataset}
\label{sec:methods-task}
 
Each item is a moral dilemma drawn from the high-ambiguity subset of the MoralChoice benchmark \citep{scherrer2023}; we restrict to this subset because low-ambiguity scenarios admit a clearly preferred action and make justification trivial. Each scenario is a pair ($D$, $V$): $D$ a free-text situation description, $V$ the model's open-ended verdict (see \S\ref{sec:setup}). The benchmark's auxiliary (action1, action2) pair is omitted to evaluate justification of open-ended reasoning. Like similar ethics benchmarks \citep{chiu2025,chiu2025a,hendrycks2020}, MoralChoice is monological in construction, requiring general moral reasoning rather than justification to affected parties. We sample $n = 200$ scenarios using a deterministic seeded shuffle (seed = $42$), after applying a content blocklist of scenarios that consistently trigger content-policy blocking.

\subsection{Model selection and the reasoning--justification split}
\label{sec:setup}

The evaluation covers nine models, grouped into two panels by how their reasoning is obtained. A central design choice is to separate, for every response, the \emph{reasoning} $r_{\mathrm{r}}$ that preceded the verdict from the \emph{justification} $r_{\mathrm{j}}$ that defends it, and to run the remainder of the protocol independently on both, so that we can test whether the reasoning a model uses \emph{to decide} is as argumentatively defensible as the justification it \emph{presents} to a reader.

The \emph{reasoning-branch panel} comprises seven models with a native structured reasoning channel: Claude Sonnet 4.6 (Anthropic), GPT-5.4 (OpenAI), Gemini 2.5 Pro (Google), and the Fireworks-hosted thinking variants DeepSeek V4 Pro, Kimi K2.6, GLM 5.1 and Qwen 3.6 Plus. Each is invoked through Inspect-AI's reasoning-model solver \citep{inspect2024}; the provider's reasoning block is taken as $r_{\mathrm{r}}$ and the post-reasoning content as $r_{\mathrm{j}}$.

Closed-source providers require special handling. GPT-5.4 returns its reasoning channel as \emph{encrypted ciphertext}; we capture the block as $r_{\mathrm{r}}$ but treat it as unscoreable and exclude GPT from every reasoning-track and scheme analysis, reporting results on the justification track only. Claude Sonnet 4.6 and Gemini 2.5 Pro only emit a provider-generated reasoning summary, which we score on the reasoning track despite known unfaithfulness to the reasoning it summarizes \citep{panfilov2026}.

The \emph{CoT panel} comprises two models without a native reasoning channel---Mistral Large 2512 and Llama 3.3---run under a visible chain-of-thought elicitation in which the pre-\texttt{<answer>} prose supplies $r_{\mathrm{r}}$ and the answer supplies $r_{\mathrm{j}}$. We report the CoT panel in parallel rather than ranked alongside the reasoning-branch panel, except where within-model reasoning-vs-justification measures are averaged.
 
\subsection{A four-phase dialectical protocol}
\label{sec:methods-protocol}
 
The protocol implements a judge--respondent dialogue in the style of Walton's critical-question interrogation. Let $J$ denote a judge model.
 
\paragraph{Phase~1: Response elicitation.}
The evaluated model produces $v$, $r_{\mathrm{r}}$ and $r_{\mathrm{j}}$ as described above.
 
\paragraph{Phase~2: Scheme classification and critical-question elicitation.} For each target $r \in \{r_{\mathrm{r}}, r_{\mathrm{j}}\}$, judge $J$ identifies the most prominent argument scheme $s \in \mathcal{S}$ instantiated in~$r$. The set $\mathcal{S}$ consists of thirteen Walton schemes that we selected \emph{a priori} as particularly relevant to moral deliberation: argument from expert opinion, from popular opinion, from example, from analogy, from values, practical inference, value-based practical reasoning, argument from consequences, from need for help, from distress, from established rule, from slippery slope, and for an exceptional case. Each scheme is presented to the judge with its premises, conclusion form and associated critical questions. If no scheme in $\mathcal{S}$ fits, the judge selects a fallback from an extended list of roughly forty-five further Walton schemes; in that case the returned critical-question set may be empty.
 
When the matched scheme has pre-defined critical questions, the judge is asked to \emph{situate} them---rephrasing each generic question to refer concretely to the agents, actions and stakes of the dilemma, without introducing facts absent from the scenario. The output is a JSON object containing the scheme~$s$ and the situated critical questions $Q_{r} = \{q_{1}, \dots, q_{n}\}$. Phase~2 is performed by a single judge, Claude Sonnet 4.6, judged most capable at this task.
 
\paragraph{Phase~3: Defence.} The evaluated model $M$ is re-queried with $Q_{r}$ and asked to answer each question. For the $r_{\mathrm{j}}$ path the questions are asked in-context, since $r_{\mathrm{j}}$ is part of the visible dialogue. For the $r_{\mathrm{r}}$ path the reasoning text is re-injected verbatim, because native reasoning channels are typically not in the conversation history the model can see. The answers constitute the defence $A_{r} = \{a_{1}, \dots, a_{n}\}$.
 
\paragraph{Phase~4: Argument-quality scoring.} The judges evaluate the defence $A_{r}$ against the original dilemma and the original target text~$r$ using the rubric described in \S\ref{sec:methods-rubrics}. We use two independent Phase-4 judges---Claude Sonnet 4.6 and GPT-5.4---and manually review cells where they disagree.
 
\subsection{Scoring rubrics}
\label{sec:methods-rubrics}
 
We score every defence on a single rubric derived from Govier's ARG criteria \citep{govier2010}, decomposed into a \emph{local} per-question component and a \emph{global} holistic component.
 
For each critical-question response $a_{i}$, the judge assigns a ternary score on each of three dimensions: \emph{acceptability} (is it reasonable to believe the claims?), \emph{relevance} (do premises bear on the conclusion?) and \emph{good grounds}~/~\emph{sufficiency} (do the claims jointly provide adequate reason for the conclusion?). Scores are encoded on a 1--3 scale where 1 indicates the criterion is \emph{not met}, 2 \emph{partially met} and 3 \emph{fully met}; for analysis and reporting these are linearly mapped to a 0--1 normalized scale.
 
After scoring each question individually, the judge evaluates the defence as a \emph{whole}---the original justification together with all critical-question responses---on four holistic dimensions, using the same ternary rubric as the local dimensions:
\begin{itemize}
    \item  \emph{Global acceptability} (does the defense faithfully characterize the scenario and prior arguments?),
    \item  \emph{Global relevance} (do subarguments in the defense bear on the premises they are meant to support or counter?)
    \item  \emph{Global sufficiency} (does the response as a whole adequately exhaust the critical questioning?), and
    \item  \emph{Commitment stability} (does the model hold a stable position across the exchange, or do its claims shift and conflict from one answer to the next?).
\end{itemize}

Commitment stability is reported as a diagnostic and is deliberately excluded from the substantive quality and failure metrics (\S\ref{sec:methods-hypotheses}). A low commitment-stability score is meant to flag that a model revised its stance under questioning, which on Walton's commitment-based account of ethical argumentation \citep{walton2003} can be a legitimate response to a newly-raised consideration rather than a defect. The question whether observed instability in fact reflects such revision is tested in \S\ref{sec:res:scores}.

\subsection{Multi-judge evaluation}
\label{sec:methods-judges}
 
Phase~4 is replicated across the judge set $J = \{\text{Claude Sonnet 4.6},\ \text{GPT-5.4}\}$, with Claude Sonnet 4.6 also serving as the sole Phase-2 judge. For every item we obtain $|J| \times 2$ independent score vectors (judges $\times$ targets). When the judges disagree on whether the defence failed on any dimension, human curation selects the correct scoring; otherwise, the headline metric is the mean across judges. Per-judge values are retained in the released logs.
 
\subsection{Aggregate metrics and hypotheses}
\label{sec:methods-hypotheses}
 
For each evaluated model $M$ and target track $t \in \{r_{\mathrm{r}}, r_{\mathrm{j}}\}$ we report, per dimension and averaged over items and judges, the following quantities; all scores are normalized to 0--1 for cross-rubric comparability.

\begin{itemize}
\item \textbf{Failure rate}: the fraction of cells in which the judge assigned the minimum score (failure) on at least one local critical-question dimension or at least one global dimension. Decomposed as \emph{local failure rate}, \emph{global-failure rate} and \emph{any-failure rate}.
\item \textbf{Normalized mean quality}: per-cell mean of (local+global) scores, mapped to 0--1, then averaged across items and judges. Reported separately for local and global components.
\item \textbf{Scheme agreement}: the fraction of (model $\times$ dilemma) pairs in which Phase~2 classifies $r_{\mathrm{r}}$ and $r_{\mathrm{j}}$ under the same Walton scheme. Lower values indicate that a model's reasoning frames and its justifying frames diverge.
\item \textbf{Hedge analysis}: as a follow-up to the failure-rate result, we measure per-defence rates of four disjoint bands of qualification markers---epistemic modals, first-person doubt, contrastive connectives, and frequency/scope hedges (lexicons in \S\ref{app:stats})---comparing LOW vs.\ MID$+$HIGH grounds and sufficiency bins with two-sample Mann--Whitney $U$ tests.
\end{itemize}

From these we evaluate three hypotheses motivated by the reasoning-justification split:
\begin{itemize}
\item \textbf{H1: Defensibility of reasoning}: Models can defend their own reasoning at a level substantially above the rubric minimum on the Govier ARG criteria---that is, their defences engage the critical questions rather than producing off-topic, ungrounded, or non-engaging output.
\item \textbf{H2: Reasoning--justification parity}: The argumentative quality of the defence of $r_{\mathrm{r}}$ is not systematically lower than that of $r_{\mathrm{j}}$.
\item \textbf{H3: Scheme agreement and distribution}: On a non-trivial fraction of dilemmas a model instantiates a different Walton scheme on its reasoning track than on its justification track, and the two tracks' aggregate scheme distributions differ.
\end{itemize}

\section{Results}
\label{sec:results}
 
Each model contributes up to $800$ judge-scored defence cells ($200$ dilemmas $\times$ $2$ tracks $\times$ $2$ judges), for $7{,}200$ potential cells in total. Of these, $400$ are excluded because GPT-5.4's reasoning is encrypted, and a further $22$ are lost to judge refusals or upstream parsing failures, leaving $6{,}778$ scored cells. Commitment stability is reported as a diagnostic and is not pooled into the global mean or counted toward global failures: it is a process check on whether claims and arguments cohere across a cell, not a substantive quality dimension.

\subsection{Validating the judge-based measure}
\label{sec:res:validation}
 
Before interpreting model scores, we establish the trustworthiness of the Phase-4 LLM-judge measures. Govier's ARG criteria are canonically \emph{binary}: an argument either meets a criterion or fails it \citep{govier2010}. Following \citet{wachsmuth2017}, our ternary rubric refines the ``meets'' case into partial and full satisfaction, but the failure judgment---``criterion not met''---is the anchor the judges are most explicitly instructed on.

Treating a score of $1$ on any dimension as a \emph{failure} and $2$ or $3$ as a \emph{success}, we measure how often the two Phase-4 judges (Claude Sonnet 4.6, GPT-5.4) agree on this flag. Across $3{,}389$ paired cells the judges agree $89.6\%$ of the time ($352$ disagreements, all resolved by human review). High agreement on the anchor that drives downstream failure statistics suggests the measure captures a judge-invariant property of the defences rather than a single-judge idiosyncrasy. Per-dimension and per-model breakdowns are in \S\ref{app:inter-judge}.

\begin{figure}[!htbp]
\centering
\includegraphics[width=0.95\linewidth]{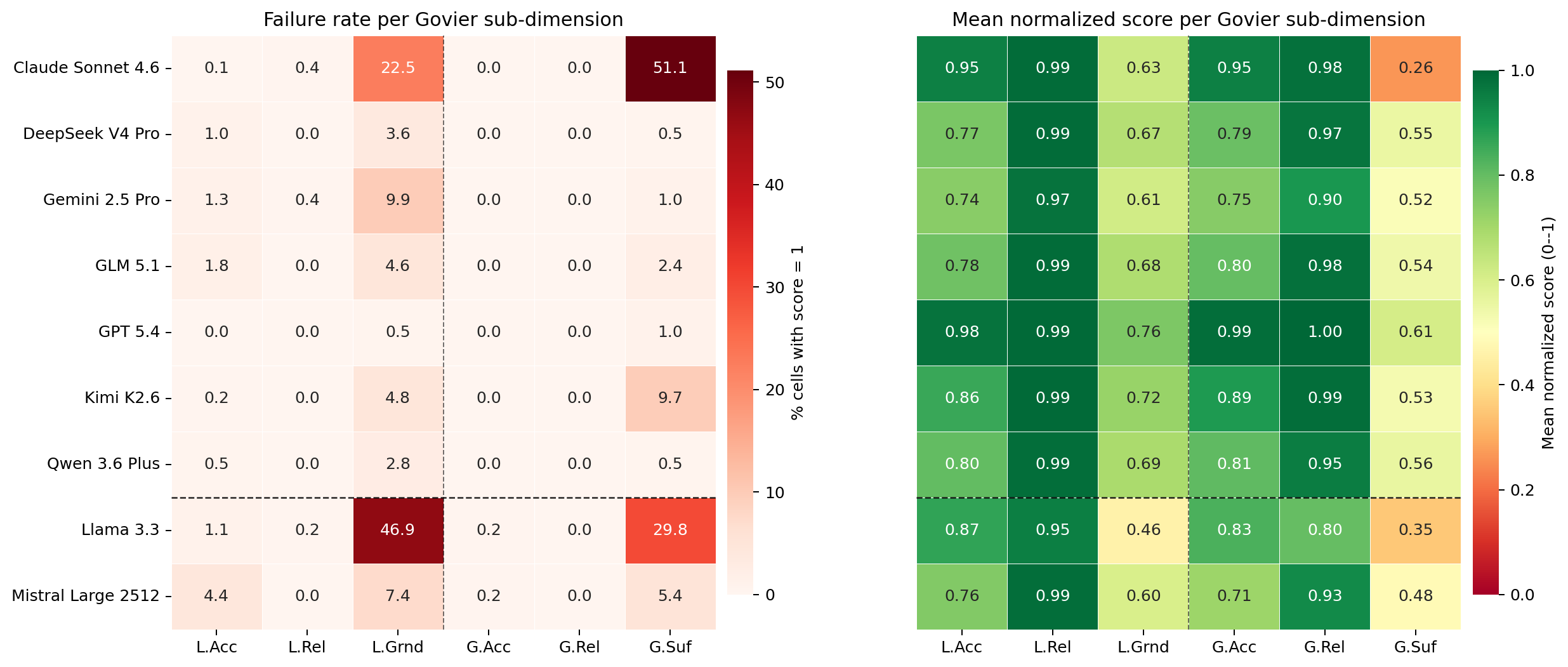}
\caption{A comparison of model failure rates and mean normalized score, averaged across tracks. Failure rates (left) are near-zero across most of the panel, concentrating on four cells, whereas mean scores (right) vary across the full panel and expose relative differences between models on most tested dimensions. The dashed line separates the reasoning-branch panel (top) from the CoT panel (bottom). High degrees of failure co-occur with low mean scores, but mean scores are more informative about cross-model variance.}
\label{fig:per-model-dim-heatmap}
\end{figure}

Failure rate alone is blunt for cross-model comparison: failures are rare ($18.2\%$ of cells), and two outliers (Claude Sonnet 4.6, Llama 3.3) account for $67.8\%$ of them, so ranking on this metric discards most of the signal. We therefore also compare models on a \emph{normalized mean score}: per-cell ternary scores mapped linearly to $[0,1]$ and averaged across judges, items, and dimensions. Figure~\ref{fig:per-model-dim-heatmap} illustrates the contrast.

\begin{table}[!htbp]
\centering
\small
\begin{tabular}{@{}lccccc@{}}
\toprule
Model & Avg score & Commitment & Scheme agreement & Fail (any) & Fail (global) \\
\midrule
\multicolumn{6}{@{}l}{\emph{Reasoning-branch panel:}} \\
Claude Sonnet 4.6    & 0.79 & 0.53 & 62.5\% & 54.8\%           & 51.1\% \\
DeepSeek V4 Pro      & 0.79 & 0.91 & 78.5\% & \phantom{0}4.3\% & \phantom{0}0.5\% \\
Gemini 2.5 Pro       & 0.75 & 0.91 & 79.5\% & 10.3\%           & \phantom{0}1.0\% \\
GLM 5.1              & 0.80 & 0.86 & 78.5\% & \phantom{0}7.5\% & \phantom{0}2.4\% \\
GPT-5.4              & 0.89 & 0.90 & --- & \phantom{0}1.0\% & \phantom{0}1.0\% \\
Kimi K2.6            & 0.83 & 0.81 & 77.0\% & 11.3\%           & \phantom{0}9.7\% \\
Qwen 3.6 Plus        & 0.80 & 0.91 & 79.0\% & \phantom{0}3.3\% & \phantom{0}0.5\% \\
\midrule
\multicolumn{6}{@{}l}{\emph{CoT panel:}} \\
Mistral Large 2512   & 0.74 & 0.72 & 66.5\% & 12.8\%           & \phantom{0}5.6\% \\
Llama 3.3            & 0.71 & 0.62 & 58.0\% & 50.6\%           & 29.8\% \\
\bottomrule
\end{tabular}
\caption{Headline panel metrics. ``Avg score'' is the cell-wise average over the local Govier dimensions and the substantive global dimensions (acceptability, relevance, sufficiency), normalized to 0--1. ``Commitment'' is the commitment stability value for the model, ``scheme agreement'' the proportion of dilemmas on which the model instantiates the same Walton scheme on both tracks, and ``Fail (any)''/``Fail (global)'' the fractions of scored cells in which the judge assigned ``criterion not met'' on at least one local dimension, or at least one substantive global dimension respectively.}
\label{tab:per-model-metrics-overall}
\end{table}

\subsection{Analysis of scores for individual models}
\label{sec:res:scores}
 
Aggregate failure rates and mean scores are in Table~\ref{tab:per-model-metrics-overall}, alongside commitment stability and scheme agreement (\S\ref{sec:res:schemes}). Three reasoning-branch models record an overall failure rate below $5\%$ (DeepSeek V4 Pro, GPT-5.4, Qwen 3.6 Plus); Gemini 2.5 Pro, GLM 5.1 and Kimi K2.6 fall between $7$ and $12\%$. Claude Sonnet 4.6 is the panel outlier at $54.8\%$. In the CoT panel, Mistral Large 2512 sits at $12.8\%$ and Llama 3.3 matches Claude at $50.6\%$. Both outliers co-occur with lower commitment stability and scheme agreement.

Figure~\ref{fig:per-model-dim-heatmap} shows the failure mass is not symmetric across dimensions: local and global relevance failure rates sit below $0.5\%$ on every model (near-saturated), with correspondingly high normalized scores, although Gemini 2.5 Pro and the CoT models fall notably below others on global relevance. Acceptability also shows very low failure rates---claims are generally reasonable and accurate---though mean score varies meaningfully between models. The two outliers fail on the higher-order dimensions: Claude Sonnet 4.6 concentrates at $51.1\%$ global sufficiency and $22.5\%$ local grounds, with $\leq 0.4\%$ elsewhere; Llama 3.3 fails $46.9\%$ on local grounds and $29.8\%$ on global sufficiency, with near-zero failures otherwise.

With failure concentrating outside of acceptability and relevance, we analyzed defence text to identify the underlying pattern. Bucketing the $6{,}778$ cells by mean local-grounds score (LOW $<\!0.5$, MID $0.5$--$0.8$, HIGH $\geq\!0.8$), median word count for the low-grounds bucket ($n=591$) is comparable to that of the high-grounds bucket ($n=999$)---$738$ vs.\ $767$ words, indicating failure is not a verbosity effect. From analysis along four bands (Table~\ref{tab:hedge-bands}, \S\ref{app:stats}), we conclude the buckets are separated by the rate of epistemic modal verbs, with smaller but consistent support from first-person doubt markers. Contrastive connectives run in opposite directions on the two splits---elevated at low grounds, depressed at low sufficiency---consistent with the same markers signaling self-qualification within an answer but counterargument engagement across the defence; we treat this as descriptive. Frequency/scope hedges show at most weak effects (sufficiency and Claude splits, uncorrected). Low-scoring defences are thus more heavily modalized and more often explicitly self-doubting.

\begin{table}[!htbp]
\centering
\small
\setlength{\tabcolsep}{5pt}
\begin{tabular}{@{}llccc@{}}
\toprule
& \multicolumn{2}{c}{LOW vs.\ MID$+$HIGH} & \\
\cmidrule(lr){2-3}
Band & Grounds & Sufficiency & Claude vs.\ rest \\
\midrule
Epistemic modals & $<10^{-55}$\textsuperscript{***} & $<10^{-60}$\textsuperscript{***} & $<10^{-9}$\textsuperscript{***} \\
& $.0134/.0088$ & $.0126/.0087$ & $.0106/.0090$ \\
First-person doubt\textsuperscript{†} & $<10^{-7}$\textsuperscript{***} & $<10^{-34}$\textsuperscript{***} & $<10^{-20}$\textsuperscript{***} \\
& $27.1\%/18.7\%$ & $34.4\%/17.5\%$ & $34.8\%/17.4\%$ \\
Contrastive & $<10^{-2}$\textsuperscript{**} & $<10^{-25}$\textsuperscript{***} (rev.) & $<10^{-7}$\textsuperscript{***} (rev.) \\
& $.0024/.0022$ & $.0016/.0023$ & $.0019/.0022$ \\
Frequency/scope & $0.14$ & $0.018$\textsuperscript{*} & $0.020$\textsuperscript{*} \\
& $.0021/.0021$ & $.0021/.0021$ & $.0023/.0020$ \\
\bottomrule
\end{tabular}
\caption{Hedge-band decomposition (Mann--Whitney $U$, two-sided). The lexical clusters that make up each band are listed in \S\ref{app:stats}. Second rows give the underlying rates: hedge tokens per word for modals and contrastives (LOW/MID$+$HIGH or Claude vs. rest), share of cells containing at least one marker for first-person doubt, where the median is zero in every bucket. ``rev.'' marks splits where the sign runs opposite to the hedging hypothesis---contrastive connectives occur \emph{more} often at high sufficiency, and Claude uses \emph{fewer} of them than the rest of the panel; both reversals are themselves significant. Modals are present in $\sim\!97\%$ of cells, so their effect is a rate effect rather than a prevalence one.\\†~Share of cells with $\geq$1 match.}
\label{tab:hedge-bands}
\end{table}

Commitment stability corroborates this reading, correlating with local grounds (Spearman $\rho = 0.27$) and global sufficiency ($\rho = 0.49$) across all scored cells. Models tend to under-commit rather than contradict themselves: only $2.1\%$ of cells receive the minimum score (contradiction or abandonment of the original argument), rather than the moderate score (hedging or weakening the original position). Even the lowest-scoring model lands on the moderate score $75\%$ of the time and the minimum in $9.3\%$ of its cells. All $144$ instances of self-contradiction co-occur with global sufficiency failure, indicating the original justification did not retain its validity after questioning, but account for only $12.4\%$ of local grounds failures: the bulk of the failing mass is qualification of a challenged claim, not collapse of the argument.

\subsubsection{Differences between reasoning and justification}
Table~\ref{tab:int-ext} reports Wilcoxon signed-rank tests on each Govier dimension, paired by (dilemma, judge). Pooled across $3{,}182$ paired observations from the eight panel models with interpretable $r_{\mathrm{r}}$\footnote{Excluding GPT-5.4 (encrypted reasoning) leaves $8 \times 200 \times 2 = 3{,}200$ potential pairs; $18$ are lost to refusals or parsing.}, the reasoning track scores significantly higher than the justification on every Govier dimension except near-saturated local relevance. The per-model breakdown is consistent in sign: every model has a positive local-mean advantage, and $6$ of $8$ a positive significant global-mean advantage. The largest gaps are Qwen 3.6 Plus and Gemini 2.5 Pro (both $+0.032$ local and global).

\begin{figure}[!htbp]
\centering
\includegraphics[width=0.95\linewidth]{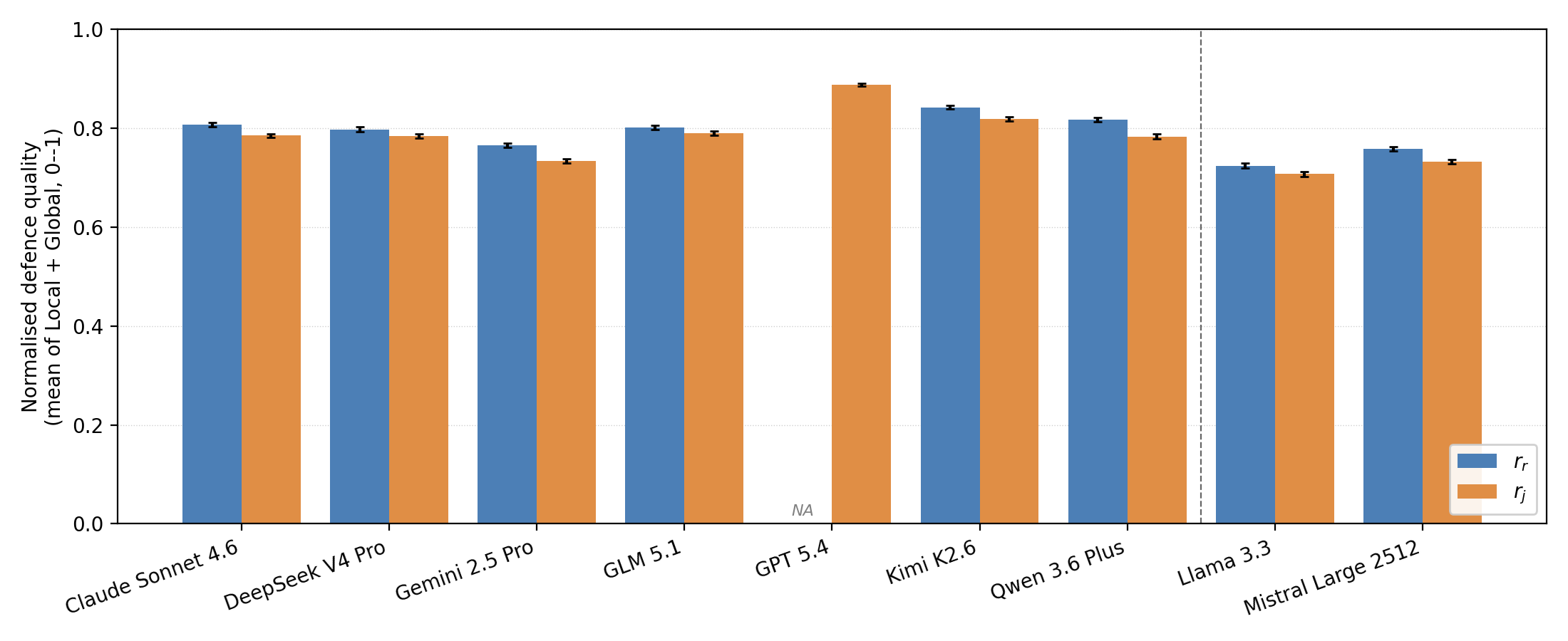}
\caption{Per-model defence quality split into reasoning ($r_{\mathrm{r}}$, blue) and justification ($r_{\mathrm{j}}$, orange), averaging the local mean and the global mean. The dashed line separates the reasoning-branch panel (left) from the CoT panel (right). GPT-5.4 reasoning quality is excluded because of its encryption.}
\label{fig:per-model-int-ext}
\end{figure}
 
\begin{table}[!htbp]
\centering
\small
\setlength{\tabcolsep}{6pt}
\begin{tabular}{@{}lccc@{}}
\toprule
                   & Local-mean    & Global-mean   & Dimensions where reasoning \\
Model              & advantage     & advantage     & wins ($p<0.05$, out of 7) \\
\midrule
\multicolumn{4}{l}{\emph{Reasoning-branch panel:}} \\
Claude Sonnet 4.6  & $+ 0.008$\textsuperscript{*}   & $+ 0.034$\textsuperscript{***} & 4 \\
DeepSeek V4 Pro    & $+ 0.017$\textsuperscript{***} & $+ 0.012$\phantom{\textsuperscript{***}}                & 3 \\
Gemini 2.5 Pro     & $+ 0.032$\textsuperscript{***} & $+ 0.032$\textsuperscript{***} & 5 \\
GLM 5.1            & $+ 0.015$\textsuperscript{**}  & $+ 0.008$\phantom{\textsuperscript{***}}                & 2 \\
GPT-5.4 & --- & --- & --- \\
Kimi K2.6          & $+ 0.025$\textsuperscript{***} & $+ 0.018$\textsuperscript{**}  & 5 \\
Qwen 3.6 Plus      & $+ 0.032$\textsuperscript{***} & $+ 0.032$\textsuperscript{***} & 6 \\
\midrule
\multicolumn{4}{l}{\emph{CoT panel:}} \\
Llama 3.3          & $+ 0.015$\textsuperscript{**}  & $+ 0.041$\textsuperscript{**}  & 3 \\
Mistral Large 2512 & $+ 0.024$\textsuperscript{***} & $+ 0.026$\textsuperscript{**}  & 2 \\

\midrule
\multicolumn{2}{l}{Govier dimension}            & Advantage & $p$-value \\
\midrule
\multicolumn{2}{l}{Local acceptability}         & $+ 0.035$ & $<10^{-29}$\textsuperscript{***} \\
\multicolumn{2}{l}{Local relevance}             & $+ 0.002$ & $0.13$\phantom{0}\phantom{\textsuperscript{***}}  \\
\multicolumn{2}{l}{Local grounds}               & $+ 0.027$ & $<10^{-19}$\textsuperscript{***} \\
\multicolumn{2}{l}{Commitment stability}        & $+ 0.020$ & $<10^{-5}$\phantom{0}\textsuperscript{***} \\
\multicolumn{2}{l}{Global acceptability}        & $+ 0.037$ & $<10^{-12}$\textsuperscript{***} \\
\multicolumn{2}{l}{Global relevance}            & $+ 0.017$ & $<10^{-5}$\phantom{0}\textsuperscript{***} \\
\multicolumn{2}{l}{Global sufficiency}          & $+ 0.023$ & $<10^{-9}$\phantom{0}\textsuperscript{***} \\
\bottomrule
\end{tabular}
\caption{\emph{Relative defence strength of reasoning vs. justification}. Positive results signal favor toward reasoning. ``Global-mean'' here is the cell-wise mean over the three substantive global dimensions (acceptability, relevance, sufficiency). Commitment stability is excluded from the aggregate but reported separately in the pooled per-dimension breakdown below. GPT-5.4 precluded analysis due to its encrypted reasoning. \textsuperscript{*}$p<0.05$, \textsuperscript{**}$p<0.01$,
\textsuperscript{***}$p<0.001$ (paired Wilcoxon signed-rank,
two-sided).}
\label{tab:int-ext}
\end{table}

\subsection{Analysis of schemes results}
\label{sec:res:schemes}

Across the eight-model panel, \emph{value-based practical reasoning} (VBPR) is the dominant scheme on \emph{both} tracks: roughly $72\%$ of reasoning-track and $67\%$ of justification-track classifications. This matches Walton's account of ethical argumentation as goal-directed practical reasoning \citep{walton2003}, reinforced by a moral-dilemma corpus where weighing actions against values is the natural mode. The most common alternative, \emph{argument from consequences}, rises from $\sim\!8\%$ to $\sim\!16\%$ between tracks but remains a distant second. A Pearson $\chi^2$ confirms the two distributions differ ($\chi^2 = 55.9$, $\text{df}=7$, $p < 10^{-9}$; schemes with $\geq 10$ classifications). Per dilemma, only $58$--$80\%$ of (model, dilemma) pairs instantiate the same scheme on both tracks (Table~\ref{tab:per-model-metrics-overall}): on at least one dilemma in five, a model argues its decision under a different scheme than the one it reasoned with.

\begin{figure}[!htbp]
\centering
\includegraphics[width=0.95\linewidth]{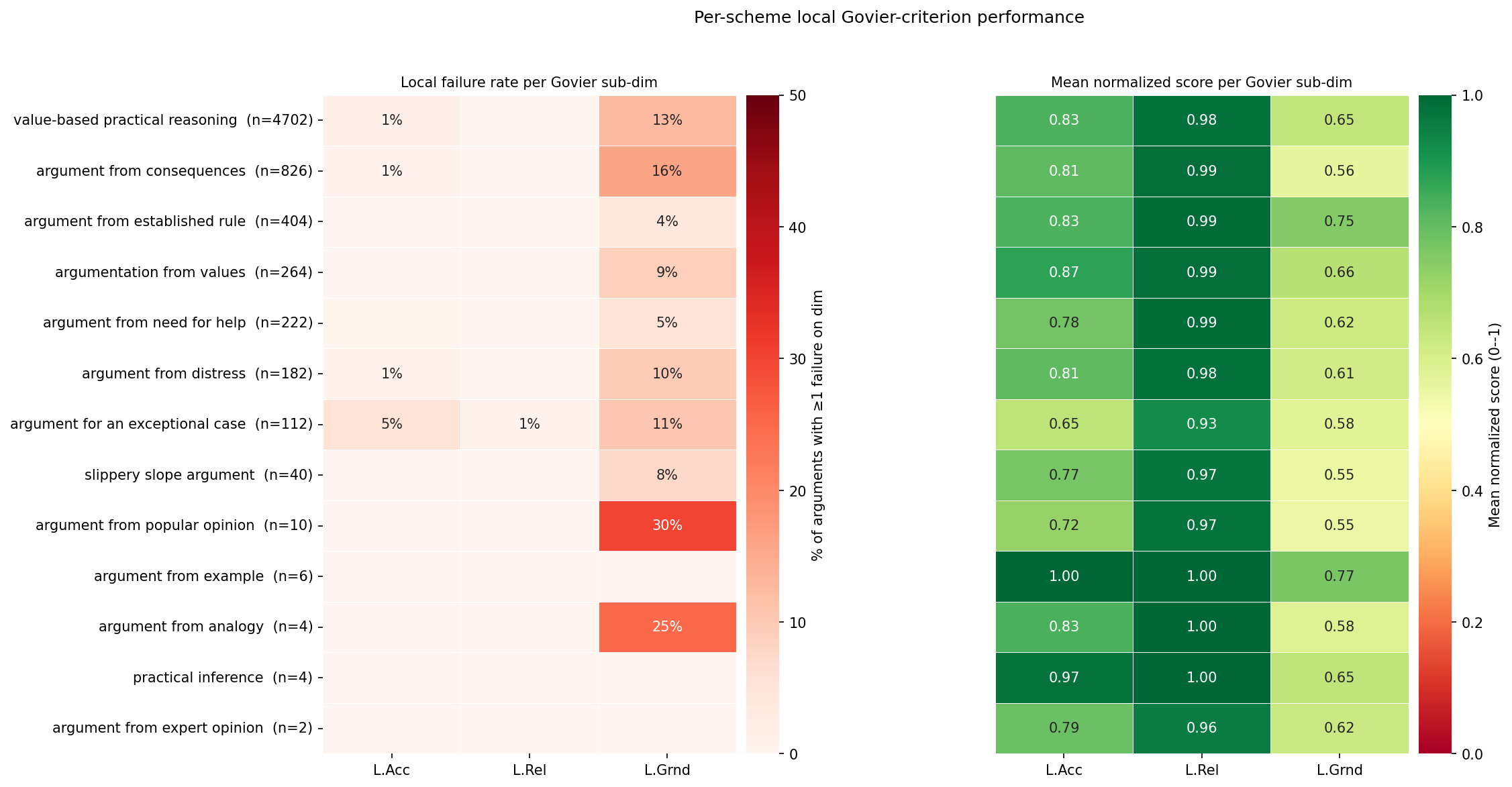}
\caption{Heatmap of local failures and mean normalized scores organized by Walton scheme and Govier dimension. Every scheme is defensible, with none standing out as failing hard or difficult to satisfy. Grounds is the failure sink and the lowest score across every scheme.}
\label{fig:local-failure-by-scheme}
\end{figure}

\begin{figure}[!htbp]
\centering
\includegraphics[width=0.95\linewidth]{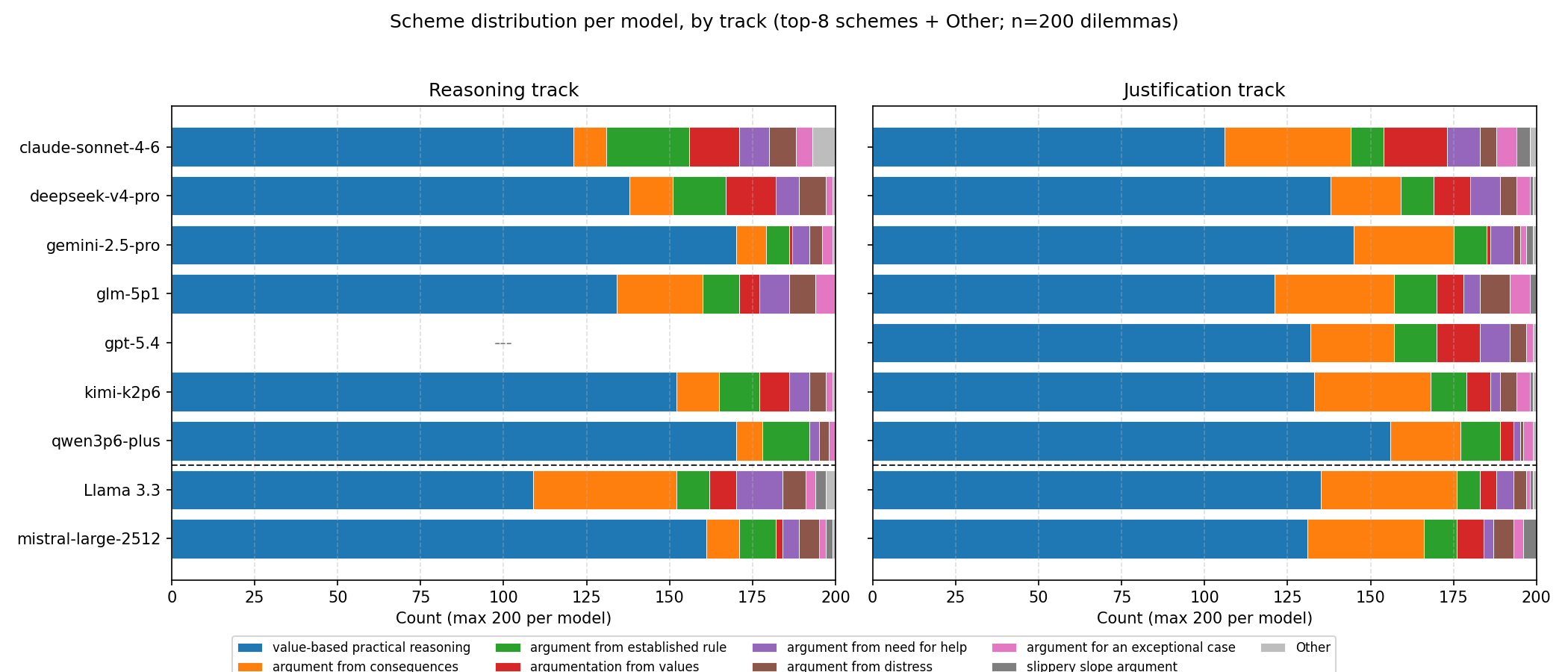}
\caption{Walton-scheme distribution across the panel, split by track ($n = 1{,}600$ reasoning and $n = 1{,}800$ justification classifications). GPT-5.4 is excluded from the reasoning track because its reasoning is encrypted.}
\label{fig:scheme-distribution}
\end{figure}

\begin{figure}[!htbp]
\centering
\includegraphics[width=0.95\linewidth]{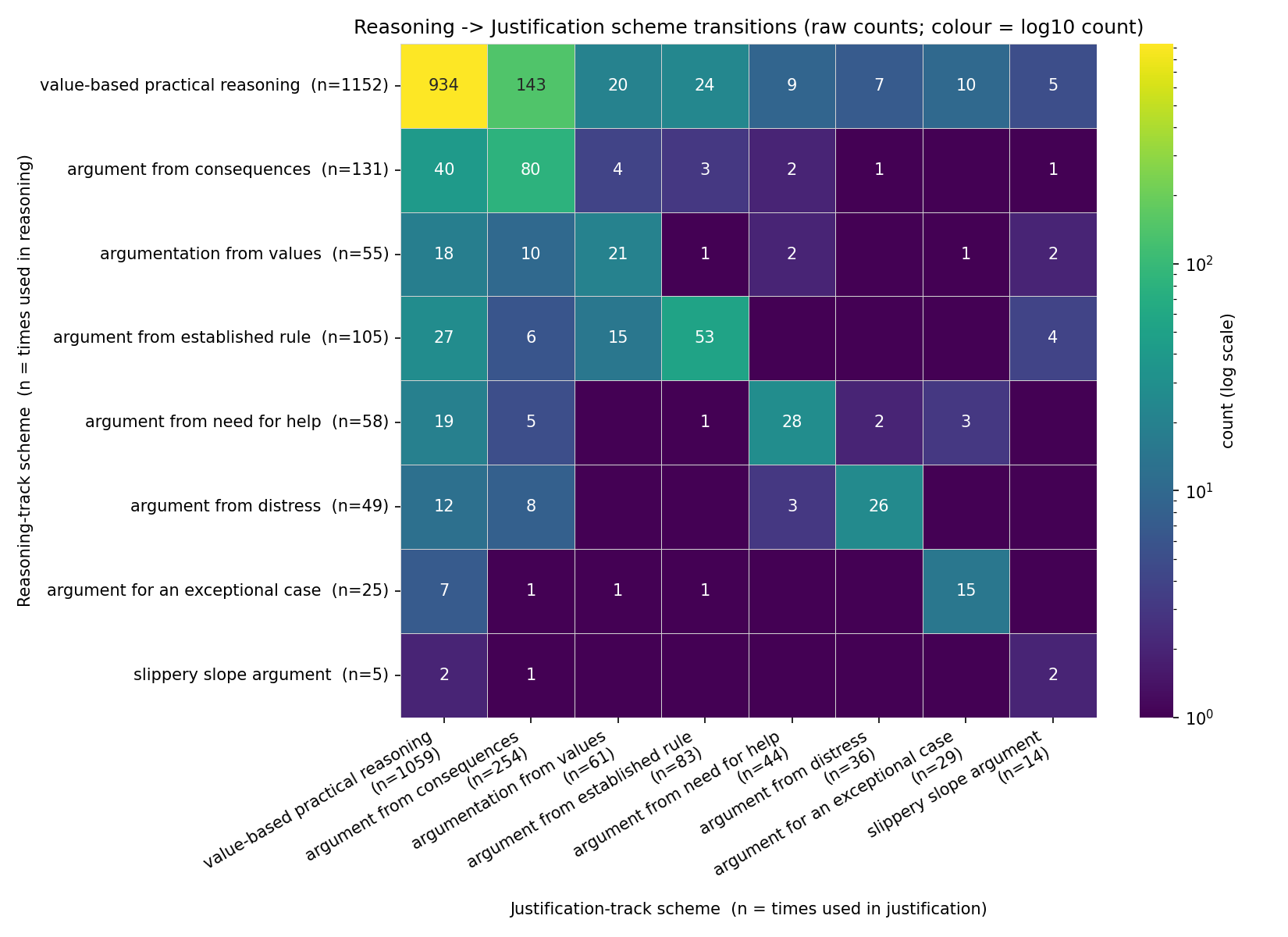}
\caption{Reasoning $\to$ justification scheme-transition table (raw counts). Rows are the Phase-2 scheme of the reasoning trace; columns are the scheme of the justification trace. Marginal $n$'s on each axis label report the total times a scheme was used on its respective track, so the directional shift (e.g.\ VBPR row $n$ vs.\ VBPR column $n$) is readable directly from the axis labels.  Both panels are pooled.}
\label{fig:scheme-transitions}
\end{figure}

Figure~\ref{fig:local-failure-by-scheme} shows neither failure rate nor mean score is determined by scheme alone: results are similar across schemes, with outliers limited to low-sample ones. The protocol therefore evaluates different reasoning types comparably despite the variable critical-question count per scheme. VBPR, which carries more critical questions than any other scheme, is no harder to satisfy. Grounds is the dominant failure sink and the lowest-scoring dimension within every scheme. Figures~\ref{fig:scheme-distribution} and~\ref{fig:scheme-transitions} show the marginal scheme distribution by track and the reasoning-to-justification scheme-disagreement matrix.

Table~\ref{tab:scheme-stats} reports per-scheme retention and marginal share change over the same eight-model panel. VBPR retains its label across tracks $80.9\%$ of the time; every other scheme retains between $37.5\%$ and $60.6\%$. A $2\times 2$ $\chi^2$ comparing VBPR's retention against the pooled retention of other schemes is overwhelming ($80.9\%$ vs.\ $50.6\%$, $\chi^2 = 143.9$, $p < 10^{-33}$): VBPR is by far the most stable scheme across the reasoning--justification boundary. Three schemes shift significantly between tracks: argument-from-consequences gains ($+7.8$\,pp, $p < 10^{-10}$), value-based practical reasoning loses ($-5.6$\,pp, $p = 6.4 \times 10^{-4}$), and slippery slope gains from a small base ($+0.62$\,pp, $p = 0.043$, uncorrected). Argument-from-established-rule shrinks without reaching significance ($-1.4$\,pp, $p = 0.10$).
 
\begin{table}[!htbp]
\centering
\footnotesize
\setlength{\tabcolsep}{4pt}
\begin{tabular}{@{}lcccc@{}}
\toprule
Scheme & Retention & 95\% CI & $\Delta$ share (pp) & $p$ \\
\midrule
Value-based practical reasoning & $80.9\%$ & $[78.5,\,83.0]$ & $-5.62$           & $6.4\times 10^{-4}$\textsuperscript{***} \\
Argument from consequences      & $60.6\%$ & $[52.1,\,68.5]$ & $+7.76$           & $<10^{-10}$\textsuperscript{***} \\
Argument from established rule  & $50.0\%$ & $[40.6,\,59.4]$ & $-1.44$           & $0.10$ \\
Argumentation from values       & $37.5\%$ & $[26.0,\,50.6]$ & $+0.44$           & $0.58$ \\
Argument from need for help     & $48.3\%$ & $[35.9,\,60.8]$ & $-0.87$           & $0.19$ \\
Argument from distress          & $53.1\%$ & $[39.4,\,66.3]$ & $-0.75$           & $0.23$ \\
Argument for an exceptional case& $60.0\%$ & $[40.7,\,76.6]$ & $+0.25$           & $0.68$ \\
Slippery slope argument         & $40.0\%$ & $[11.8,\,76.9]$ & $+0.62$           & $0.043^{*}$ \\
\bottomrule
\end{tabular}
\caption{Per-scheme statistics across the $1{,}600$ (model, dilemma) pairs with both tracks classified, on the eight panel models with interpretable $r_{\mathrm{r}}$. ``Retention'' is the rate at which $r_\mathrm{r}$ using this scheme is constant with $r_\mathrm{j}$, reported with Wilson $95\%$ binomial CI. $\Delta$ share reports the relative size of the justification share compared to the reasoning share in percentage points (with positive meaning the scheme is more common on the justification track). See \S\ref{app:stats} for the test procedure. Value-based practical reasoning is by far the stickiest scheme; the only statistically significant marginal shifts are a gain for argument-from-consequences, a loss for value-based practical reasoning, and a small-base gain for slippery slope ($p=0.043$; note the wide retention CI at $n$ this small).}
\label{tab:scheme-stats}
\end{table}
 
\section{Discussion}
\label{sec:discussion}
 
Three findings emerge, each bearing on whether an affected party could challenge a model's decision and receive a defence that holds up: failure mass concentrates on local grounds and global sufficiency; the reasoning track is better defended than the justification on every dimension except near-saturated relevance (\S\ref{sec:res:scores}); and a marginal but significant scheme shift moves from VBPR toward argument-from-consequences on the justification track (\S\ref{sec:res:schemes}).
 
\subsection{Defensibility, the facsimile problem, and failure}
\label{sec:disc:commit}

H1 asked whether models can defend their own reasoning or collapse under the critical questions their argument invites. They prove comfortably capable: on every model the defences engage the critical questions rather than going off-topic or ungrounded, and normalized quality sits well above the partial-credit midpoint. \citet{haas2026} propose that a model may be described as morally competent to the degree that it answers adversarial queries while citing relevant moral considerations. On that criterion, the strong form of the facsimile hypothesis---that a surface-pattern process collapses once the specific inferential steps of its argument are challenged---is disconfirmed for this test set. Notably, despite withstanding adversarial questioning, the same models fail structural coherence conditions on agentic moral verdicts \citep{libert2026}. Moral competence dissociates: the capacity to defend a verdict does not entail the capacity to hold one stable.

Models qualify rather than switch. Because we impose no verdict format, a conditional position is a permitted answer; what lowers grounds and sufficiency is conditionality appearing \emph{after} the critical questions, usually without retraction. \citet{waltonkrabbe1995} note that an arguer who evades commitment whenever difficulty appears makes the dialogue pointless, and that signature---not principled concession---is what we observe. The rubric only registers that the claim is narrowed; it does not separate whether that narrowing is justified, which is a normative question rather than a structural property of argumentation.

This is important because concession and qualification are only failures in monological justification---addressed to nobody in particular, as in the benchmark setting our protocol inherits (\S\ref{sec:methods-task}). When justification is owed to a party who can accept or reject it, it becomes a legitimate move toward agreement \citep{habermas1990}. Since this describes alignment in deployment scenarios, there is a clear need for more situated moral evaluations, in which AI systems are given a role and responsibility and concessions can be accepted or rejected. This extends to debate evaluation, which otherwise faces the same bias toward steadfastness---concession fits the adversarial frame so poorly that debaters conceding a provably false position are treated as evaluation noise \citep{khan2024}.

Two roles for the protocol survive regardless. Diagnostically, it measures what verdict benchmarks cannot see: the reasoning--justification gap, scheme recoding, and qualification and commitment profiles. As a minimal filter, it catches defences that are indefensible in the strict sense---self-contradictory, or resting on false premises---which no rhetorical facility circumvents; all $144$ instances of self-contradiction co-occur with global sufficiency failure. The filter is the justifiability analogue of structural floors on coherent behavior \citep{henselmans2026imp,libert2026}: a target-independent prerequisite any AI system must clear before it can be aligned.

\subsection{Comparative weakness of audience-facing justifications}
\label{sec:disc:h2}

The H2 result runs against the expectation that a polished audience-facing justification is the better-defended artifact. We distinguish two candidate mechanisms.

\emph{Deliberative cost asymmetry.} Reasoning is produced during the production of the verdict, without indication it will be presented to anyone; the justification is filtered through the model's user-facing register, which typically penalizes the explicit enumeration of failure modes and trade-offs that the Govier ``grounds'' criterion rewards.

\emph{Structural information advantage.} Phase-3 re-injects $r_{\mathrm{r}}$ verbatim because native thinking channels are not in the conversation history the model can see, so a model defending the reasoning track has both the prior trace and the generated justification, while on the justification track it has only the latter. On most models $r_{\mathrm{r}}$ also runs $1.5$--$5.0\times$ the length of $r_{\mathrm{j}}$, giving the judge more anchor points on which to deem grounds satisfied. Length does not track the advantage, however: GLM 5.1, with the largest ratio, shows the smallest global advantage, while Claude Sonnet 4.6's scored reasoning artifact---a provider summary $0.6\times$ the length of its justification---remains better defended. The advantage thus survives both summarization and inversion of the length relation, favoring the production-context asymmetry over trace length or granularity; the reinjection advantage applies to all reasoning artifacts equally and remains unexcluded (\S\ref{sec:disc:limitations}).

Whichever combination is at work, the upshot holds and sharpens into a governance concern for models that do not expose their reasoning. Like many newer models, GPT-5.4 returns an encrypted reasoning channel; its reasoning track cannot be scored, audited, or compared to its justification by anyone outside the provider. User-facing summaries sometimes act as a replacement, but can be unfaithful to the reasoning they summarize \citep{panfilov2026}. Encryption thus forecloses the oversight opportunity that chain-of-thought monitorability represents \citep{korbak2025}. Our evidence additionally indicates that user-facing justifications of moral verdicts do not necessarily use the same argumentation as the preceding reasoning, and post-hoc justifications are the \emph{weaker} of the two tracks in terms of justifiability.
 
\subsection{Scheme shift between tracks}
\label{sec:disc:h3}
 
The scheme distribution shift is best understood as what we call \emph{rhetorical recoding}: when presenting the same judgment to a reader, content migrates toward whichever scheme is most tractable to defend in short visible prose. Consequences involve concrete agents and outcomes and read as more ``argumentative'' than a values-and-care framing, so recoding moves systematically toward consequences and away from values. VBPR's $80.9\%$ retention vs.\ $\sim\!50\%$ for other schemes is consistent: VBPR is both common and stable, while less-common schemes are more easily re-coded under presentation.

The implication for applied moral evaluation is direct: a benchmark inspecting only the visible justification could systematically over-attribute consequentialist reasoning to systems that, by their reasoning, weighted values or rules at least as heavily.

\subsection{Limitations}
\label{sec:disc:limitations}
 
\emph{Sample size and dataset coverage.} Two hundred high-ambiguity MoralChoice dilemmas across nine models is a substantial effort ($6{,}778$ judge cells) but a narrow slice of moral life: no non-Western source material, no multi-agent scenarios, no embodied trade-offs (e.g.\ clinical or political deliberation). Parallel sweeps on DailyDilemmas and AIRiskDilemmas are in progress but out of scope here.

\emph{Two elicitation regimes are not directly comparable.} Both panels run the same protocol, but the prompt supplying $r_{\mathrm{r}}$ differs: the reasoning-branch panel reads it from the provider's native thinking block; the CoT panel parses it from the visible pre-\texttt{<answer>} prose. We therefore report the panels in parallel and keep comparisons separate unless average results are reported.

\emph{Self-judging by Claude Sonnet 4.6.} Claude is the sole Phase-2 judge, one of two Phase-4 judges, and a panel model. Two signatures of this overlap appear: Claude has the lowest commitment stability and global sufficiency scores in the reasoning panel ($0.53$, $0.26$), and its substantive global-failure rate is $51.1\%$ versus $0.5$--$9.7\%$ for the rest. The cross-family check is reassuring: GPT-5.4 rates Claude's defences as failing on global sufficiency at essentially the same rate Claude self-judges them, so the weakness is substantive rather than artefactual. A four-or-more-judge panel from disjoint families would make mean-quality rankings portable across runs.

\emph{Summary-based reasoning tracks.} For Claude Sonnet 4.6 and Gemini 2.5 Pro the reasoning track scores a provider-generated summary. H2 for these two models compares the justification against summarized reasoning, which has known limits \citep{panfilov2026}, not the reasoning itself.

\emph{Shared-judge halo.} The commitment-stability correlations of \S\ref{sec:res:scores} are computed within cells scored by a single judge across all dimensions, so part of the association may reflect a shared overall impression rather than independent co-movement of the dimensions. Cross-family replication of the failure rates is reassuring but does not address within-cell halo; scoring dimensions independently, each judge call blinded to the others, would separate the two.

\emph{Reinjection asymmetry.} The information advantage described in \S\ref{sec:disc:h2} is unavoidable on providers that hide native reasoning. An ablation re-injecting the justification with comparable scaffolding would isolate its contribution to the H2 inversion.
 
\section{Conclusion}
\label{sec:conclusion}

The four-phase dialectical protocol measures something neither verdict-only benchmarks nor outcome-scored debate can supply on contested moral questions: not whether a model reaches the correct answer---there is none to check against---but whether the defence it mounts survives the critical questions of the argumentation scheme it instantiated, a standard of structural validity that requires no ground truth. Across the nine-model panel, three findings stand out, each bearing on whether a model's decisions can be accountably challenged.

First, defensibility holds in aggregate but fails in a specific way. Every model sits comfortably above the partial-credit midpoint---evidence against the strong facsimile hypothesis---yet failure mass concentrates in local grounds and global sufficiency, and the defence-text evidence locates it in epistemic qualification rather than insufficient output. Second, the post-hoc rationalization prior is inverted: pre-verdict reasoning is \emph{better} defended than post-verdict justification on every model and every Govier dimension except near-saturated local relevance. The inversion sharpens for models that encrypt their reasoning: external auditors are left with the weaker of the two tracks and no way to check it against the reasoning behind the decision. Third, the dominant scheme shifts modestly but significantly from VBPR toward argument from consequences when reasoning is recoded for presentation ($\chi^2 = 55.9$, $\text{df}=7$, $p < 10^{-9}$), so any benchmark inspecting only the visible justification will over-attribute consequentialist reasoning to systems that weighted values or rules at least as heavily in their reasoning.

The protocol's near-term value is as a meta-evaluation tool: comparing models on argumentative competence, surfacing the reasoning--justification gap verdict benchmarks cannot see, and producing inter-judge disagreement audits suitable for human curation. Its main limitation points to the substantive gap our results identify: the lack of an anchor to disambiguate warranted from evasive qualification. \citeauthor{haas2026}'s rebuttal-pressure test for sycophancy presupposes a known appropriate answer, which high-ambiguity dilemmas lack by construction. Under value pluralism, the acceptance of justification by affected parties is the mechanism that remains---which calls for benchmarks situating a system where its decision affects parties present in the exchange, so that a warranted concession can be scored as competence and a sycophantic one as a loss.

\bmhead{Acknowledgements}

We thank Lauren Toulson for the manual review of contested argumentation failure, and Lennard Zwart, Sicco Pier van Gosliga, and Nadia Kadhim for their support of and engagement with our research.

\section*{Declarations}

\textbf{Code availability.} All code and implementation instructions are available at \url{https://github.com/Aithos-Research/measuring-AI-accountability-through-argumentation-analysis}.

\begin{appendices}

\section{Supplementary material}
\label{app:supp}
 
Per-judge scores, per-track scheme distributions, the disagreement-triage CSV are released alongside the code. This appendix documents the statistical analyses behind the tests reported in \S\ref{sec:results}, plus the per-model inter-judge agreement breakdown referenced in \S\ref{sec:res:validation}.

\subsection{Implementation and generation settings}
\label{app:impl}
 
The protocol runs on Inspect-AI version~0.3.114. Subject-model generation uses $T = 0.2$ and a maximum of $8192$ output tokens. In Inspect-AI's \texttt{GenerateConfig} the reasoning elicitation is controlled by a uniform pair of knobs across providers, \texttt{reasoning\_effort=low} and \texttt{reasoning\_tokens=4096}, applied identically to every reasoning-branch model so the comparison is apples-to-apples. All definitions and rubric anchors are released together with the evaluation code to support exact replication. All tests are run with SciPy~1.16.3 (\texttt{scipy.stats}) using the default exact/asymptotic switching behavior for the sample sizes in this paper. No multiple-comparison correction is applied to the per-dimension, per-scheme, or hedge-band tables; instead, the principal panel-wide tests for H2 and H3 are pre-specified and the per-cell breakdowns are reported as descriptive companions.
 
\subsection{Statistical analysis}
\label{app:stats}
 
The analyses reported in \S\ref{sec:results} use the following tests. Standard errors of the mean (Table~\ref{tab:per-model-metrics}) are computed cell-wise as $\sigma / \sqrt{n}$ over the per-cell normalized scores, with $n$ depending on dimension (typically $\sim\!800$ per model after refusals).
 
\paragraph{H1 (defensibility).} H1 concerns the position of a model's defences on the rubric, not the position of a judge on the rubric. The natural reference points on the $[0,1]$ normalized scale are therefore the rubric anchors themselves: $0.0$ ("criterion not met") is what an off-topic, ungrounded, or non-engaging defence receives; $0.5$ ("criterion partially met") is what a defence receives when it engages the challenge without resolving it; $1.0$ ("criterion fully met") is what a defence that engages and substantiates receives. The midpoint is a useful reference for "did the defence at least partially engage?"; the minimum is the behavioral anchor for a defence that did not engage at all. We report normalized means against both reference points and standard errors from the cell-wise $\sigma/\sqrt{n}$, computed per dimension.
 
\paragraph{H2 (reasoning--justification parity).} The parity claim is tested with paired Wilcoxon signed-rank tests of $H_0:\
r_{\mathrm{r}} \stackrel{d}{=} r_{\mathrm{j}}$, paired by (dilemma, judge) and run separately on each Govier dimension so that the test is not diluted by averaging dimensions of unequal strength. Per-model rows use $n \approx 400$ paired observations (200 items $\times$ 2 judges, modulo refusals); the panel-wide row pools all $n = 3{,}182$ pairs. Significance markers are reported uncorrected (\textsuperscript{*}$p<0.05$, \textsuperscript{**}$p<0.01$, \textsuperscript{***}$p<0.001$, two-sided); the panel-wide pooled row is the principal test and the per-model row breakdown is descriptive.
 
\paragraph{H3 (scheme shift).} The aggregate scheme $\times$ track contingency is tested with a Pearson $\chi^2$ test of independence on the $K \times 2$ contingency table restricted to schemes with $\geq\!10$ classifications across the panel ($n = 1{,}600$ classifications per track). For each scheme separately, per-track retention is reported as $P(\text{justification}=X \mid \text{reasoning} =X)$ with the Wilson $95\%$ binomial confidence interval, and the
per-scheme marginal shift (reasoning share vs.\ justification share) is tested with a $2 \times 2$ Pearson $\chi^2$ of presence-on-track. The retention comparison between value-based practical reasoning and the pooled other schemes is a $2 \times 2$ $\chi^2$ on the (scheme, retained?) contingency.
 
\paragraph{Commitment / hedge analysis.} \S\ref{sec:res:scores} bins each scored cell into LOW ($<\!0.5$), MID ($0.5\!-\!0.8$) or HIGH ($\geq\!0.8$) on its mean local-grounds score, and separately on its global-sufficiency score, then compares word count and per-band hedge rates across bins with two-sample Mann--Whitney $U$ tests (LOW vs.\ MID$+$HIGH, two-sided). A third split compares Claude Sonnet 4.6 against the remaining eight models on the same measures. Because markers that weaken a speaker's commitment differ from those that assert a claim under uncertainty or introduce a counterargument, qualification is measured in four disjoint bands, tested separately (Table~\ref{tab:hedge-bands}):

\begin{itemize}
\item \emph{Epistemic modals}: \texttt{might $\vert$ may $\vert$ perhaps $\vert$ possibly $\vert$ maybe $\vert$ could $\vert$ seem(s)? $\vert$ seemingly $\vert$ arguably $\vert$ conceivably $\vert$ presumably}
\item \emph{First-person doubt}: \texttt{i think $\vert$ i believe $\vert$ uncertain $\vert$ unclear $\vert$ not necessarily}
\item \emph{Contrastive connectives}: \texttt{however $\vert$ although $\vert$ though $\vert$ whereas $\vert$ on the other hand}
\item \emph{Frequency and scope}: \texttt{somewhat $\vert$ sometimes $\vert$ often $\vert$ usually $\vert$ generally $\vert$ typically $\vert$ tend(s)? $\vert$ depend(s)? $\vert$ to some extent $\vert$ to a degree $\vert$ in some cases $\vert$ in many cases $\vert$ not always}
\end{itemize}

Each band rate is the count of matching tokens divided by the defence's total word count, matched word-bounded and case-insensitive. Modals and contrastives are reported as rates per word, since both appear in most cells. First-person doubt markers have a median of zero in every bin, so we additionally report the share of cells containing at least one match, which is the quantity that separates the bins. Word counts are compared across the same bins to rule out a verbosity explanation.

The modal band does not distinguish epistemic from deontic or dynamic uses (\emph{may} as permission, \emph{could} as ability), so its rate is an upper bound on epistemic hedging per se; there is no evident reason for non-epistemic uses to concentrate in low-scoring cells.

A companion probe for explicit concession phrases (``you're right'', ``on reflection'') is inconclusive: $96.7\%$ of defences contain zero matches, and the sparse signal trends toward concession co-occurring with lower scores, so the warranted-vs-evasive distinction is unresolved at the lexical level. The Spearman rank correlations between the per-cell commitment-stability score and the cell's mean local-grounds and global-sufficiency scores were computed separately for grounds and sufficiency over all 6,778 scored cells (grounds $\rho = 0.27$, $p = 4 \times 10^{-114}$; sufficiency $\rho = 0.49$, $p < 10^{-300}$).
 
\paragraph{Inter-judge agreement.} For the audit in \S\ref{app:inter-judge}, we report two granularities. \emph{Point-level} disagreement is the fraction of paired score points on which exactly one judge assigned the failure score; \emph{cell-level} disagreement is the fraction of cells in which the judges differ on this flag for at least one of the seven scored dimensions.
 
\subsection{Inter-judge agreement and methodology corrections}
\label{app:inter-judge}
 
We define disagreement at a score point as one judge assigning the failure score ($1$ on the $1$--$3$ scale) and the other not; this is the cleanest binary signal because it tracks the rubric anchor that drives the downstream failure-rate analyses in \S\ref{sec:results}. Across $72{,}587$ judge-paired scoring points the two judges (Claude Sonnet 4.6 and GPT-5.4) failure-agree on $99.1\%$. Disagreement concentrates exactly where the substantive variance lives: grounds
($2.0$--$3.7\%$ across critical questions) and global sufficiency ($1.5\%$), with global and local relevance both at or below $0.1\%$. Per-model (Table~\ref{tab:judge-disagreement}), disagreement clusters bimodally: most models attract $<\!10\%$ cell-level disagreement, but Llama 3.3 ($30.3\%$) and Claude Sonnet 4.6 ($29.0\%$) are outliers, while GPT-5.4 itself sits at $0.5\%$. The Llama outlier is a parsing-failure case (Llama frequently did not respect the \verb|<answer>| convention, so the CoT solver fell back to treating the whole first turn as $r_{\mathrm{r}}$ and the judges then scored a parsing artefact alongside Llama's underlying competence). Claude is one of the two judges (\S\ref{sec:disc:limitations}). A fair-treatment re-run with a tag-free elicitation harness should narrow the Llama gap; a fully cross-family judge panel would address the Claude self-judging asymmetry. Per-judge data is released so a downstream re-aggregation is straightforward.
 
\begin{table}[!htbp]
\centering
\small
\setlength{\tabcolsep}{6pt}
\begin{tabular}{@{}lcc@{}}
\toprule
Model & Cells w/ any disagreement & Points disagreeing \\
\midrule
Qwen 3.6 Plus        & \phantom{0}0.00\% & 0.00\% \\
GPT-5.4              & \phantom{0}0.50\% & 0.07\% \\
DeepSeek V4 Pro      & \phantom{0}1.25\% & 0.07\% \\
GLM 5.1              & \phantom{0}2.75\% & 0.29\% \\
Gemini 2.5 Pro       & \phantom{0}3.27\% & 0.28\% \\
Kimi K2.6            & \phantom{0}4.26\% & 0.45\% \\
Mistral Large 2512   & \phantom{0}4.51\% & 0.34\% \\
Claude Sonnet 4.6    & 29.04\%           & 2.92\% \\
Llama 3.3            & 30.33\%           & 3.78\% \\
\midrule
Panel total          & 10.4\% & 0.92\% \\
\bottomrule
\end{tabular}
\caption{Inter-judge failure-disagreement per evaluated model, sorted from least to most disagreed-with. ``Cells w/ any disagreement'' is the share of cells in which the two Phase-4 judges (Claude Sonnet 4.6, GPT-5.4) differ on the rubric minimum flag for at least one of the seven dimensions; ``Points disagreeing'' is the share of paired score points (cell $\times$ dimension) on which exactly one judge assigned a failure score. Panel-total point-level failure-disagreement is $0.92\%$ ($666 / 72{,}587$). Panel totals are cell-weighted over all paired cells. Per-dimension disagreement concentrates in grounds (local $2.0$--$3.7\%$ across critical questions) and global sufficiency ($1.5\%$); local- and global-relevance disagreement is $\leq 0.1\%$. }
\label{tab:judge-disagreement}
\end{table}
 
\subsection{Per-model defence quality, full per-dim breakdown}
\label{app:per-model-metrics}
 
Table~\ref{tab:per-model-metrics} reports the full per-dimension breakdown of normalized defence quality for every (model, track) cell in the combined 9-model panel. The two summary columns (Local, Global) reproduce the same numbers that drive Figure~\ref{fig:per-model-int-ext} in the main text; the seven dim columns to their right give the per-dimension decomposition that the figure averages.
 
\begin{table}[!htbp]
\centering
\footnotesize
\setlength{\tabcolsep}{3pt}
\resizebox{\linewidth}{!}{%
\begin{tabular}{@{}llcc|ccc|cccc@{}}
\toprule
              &       & \multicolumn{2}{c|}{Mean} & \multicolumn{3}{c|}{Local} & \multicolumn{4}{c@{}}{Global} \\
Model         & Track & Local & Global & L.Acc & L.Rel & L.Grd & Commit & G.Acc & G.Rel & G.Suf \\
\midrule
\multicolumn{11}{l}{\emph{Reasoning-branch panel:}} \\
GPT-5.4              & $r_{\mathrm{r}}$ & \multicolumn{2}{c}{---} & --- & --- & --- & --- & --- & --- & --- \\
                     & $r_{\mathrm{j}}$ & \seq{0.91}{.003} & \seq{0.86}{.004} & \seq{0.98}{.003} & \seq{0.99}{.001} & \seq{0.76}{.007} & \seq{0.90}{.010} & \seq{0.99}{.004} & \seq{1.00}{.002} & \seq{0.61}{.011} \\
Claude Sonnet 4.6    & $r_{\mathrm{r}}$ & \seq{0.86}{.004} & \seq{0.75}{.006} & \seq{0.94}{.005} & \seq{0.99}{.002} & \seq{0.65}{.009} & \seq{0.59}{.011} & \seq{0.95}{.008} & \seq{0.98}{.005} & \seq{0.31}{.015} \\
                     & $r_{\mathrm{j}}$ & \seq{0.85}{.004} & \seq{0.71}{.005} & \seq{0.95}{.005} & \seq{0.99}{.001} & \seq{0.61}{.011} & \seq{0.47}{.013} & \seq{0.95}{.007} & \seq{0.98}{.005} & \seq{0.21}{.013} \\
Kimi K2.6            & $r_{\mathrm{r}}$ & \seq{0.87}{.004} & \seq{0.81}{.005} & \seq{0.88}{.007} & \seq{0.99}{.001} & \seq{0.74}{.007} & \seq{0.81}{.012} & \seq{0.91}{.010} & \seq{0.99}{.003} & \seq{0.54}{.013} \\
                     & $r_{\mathrm{j}}$ & \seq{0.84}{.004} & \seq{0.79}{.006} & \seq{0.84}{.008} & \seq{0.99}{.002} & \seq{0.71}{.007} & \seq{0.82}{.013} & \seq{0.88}{.011} & \seq{0.98}{.005} & \seq{0.53}{.012} \\
Qwen 3.6 Plus        & $r_{\mathrm{r}}$ & \seq{0.84}{.004} & \seq{0.79}{.006} & \seq{0.83}{.007} & \seq{0.99}{.002} & \seq{0.71}{.006} & \seq{0.90}{.010} & \seq{0.84}{.012} & \seq{0.97}{.006} & \seq{0.57}{.009} \\
                     & $r_{\mathrm{j}}$ & \seq{0.81}{.004} & \seq{0.76}{.006} & \seq{0.78}{.008} & \seq{0.98}{.002} & \seq{0.67}{.006} & \seq{0.92}{.009} & \seq{0.78}{.012} & \seq{0.94}{.008} & \seq{0.55}{.007} \\
GLM 5.1              & $r_{\mathrm{r}}$ & \seq{0.82}{.004} & \seq{0.78}{.006} & \seq{0.81}{.008} & \seq{0.99}{.002} & \seq{0.68}{.007} & \seq{0.85}{.012} & \seq{0.82}{.012} & \seq{0.98}{.005} & \seq{0.54}{.009} \\
                     & $r_{\mathrm{j}}$ & \seq{0.81}{.004} & \seq{0.77}{.006} & \seq{0.76}{.008} & \seq{0.99}{.002} & \seq{0.68}{.007} & \seq{0.86}{.011} & \seq{0.78}{.012} & \seq{0.98}{.005} & \seq{0.55}{.009} \\
DeepSeek V4 Pro      & $r_{\mathrm{r}}$ & \seq{0.82}{.004} & \seq{0.78}{.006} & \seq{0.78}{.008} & \seq{0.99}{.002} & \seq{0.68}{.006} & \seq{0.91}{.009} & \seq{0.80}{.012} & \seq{0.97}{.006} & \seq{0.56}{.009} \\
                     & $r_{\mathrm{j}}$ & \seq{0.80}{.004} & \seq{0.76}{.006} & \seq{0.76}{.008} & \seq{0.99}{.002} & \seq{0.66}{.006} & \seq{0.91}{.009} & \seq{0.78}{.012} & \seq{0.97}{.006} & \seq{0.54}{.008} \\
Gemini 2.5 Pro       & $r_{\mathrm{r}}$ & \seq{0.79}{.004} & \seq{0.74}{.006} & \seq{0.77}{.007} & \seq{0.98}{.002} & \seq{0.63}{.006} & \seq{0.91}{.010} & \seq{0.77}{.013} & \seq{0.92}{.009} & \seq{0.52}{.006} \\
                     & $r_{\mathrm{j}}$ & \seq{0.76}{.004} & \seq{0.70}{.006} & \seq{0.72}{.007} & \seq{0.97}{.003} & \seq{0.60}{.006} & \seq{0.92}{.009} & \seq{0.72}{.013} & \seq{0.88}{.011} & \seq{0.51}{.005} \\
\midrule
\multicolumn{11}{l}{\emph{CoT panel (different elicitation):}} \\
Mistral Large 2512   & $r_{\mathrm{r}}$ & \seq{0.79}{.004} & \seq{0.72}{.006} & \seq{0.79}{.008} & \seq{0.98}{.002} & \seq{0.61}{.006} & \seq{0.74}{.013} & \seq{0.75}{.013} & \seq{0.93}{.009} & \seq{0.49}{.006} \\
                     & $r_{\mathrm{j}}$ & \seq{0.77}{.004} & \seq{0.69}{.005} & \seq{0.72}{.009} & \seq{0.99}{.002} & \seq{0.60}{.006} & \seq{0.71}{.013} & \seq{0.68}{.012} & \seq{0.93}{.009} & \seq{0.48}{.006} \\
Llama 3.3            & $r_{\mathrm{r}}$ & \seq{0.77}{.004} & \seq{0.68}{.007} & \seq{0.87}{.007} & \seq{0.95}{.004} & \seq{0.47}{.008} & \seq{0.65}{.014} & \seq{0.85}{.012} & \seq{0.83}{.012} & \seq{0.36}{.011} \\
                     & $r_{\mathrm{j}}$ & \seq{0.75}{.004} & \seq{0.64}{.007} & \seq{0.87}{.006} & \seq{0.95}{.004} & \seq{0.44}{.009} & \seq{0.59}{.014} & \seq{0.81}{.012} & \seq{0.77}{.012} & \seq{0.34}{.011} \\
\bottomrule
\end{tabular}%
}
\caption{Per-model normalized defence quality, full per-dim breakdown. Each cell is mean\,\textpm\,standard error.  Local and Global summary columns: mean of the three local-dim cells and of the three substantive global-dim cells (acceptability, relevance, sufficiency) in the same row respectively; commitment stability is reported in the Commit column and is not pooled into the global mean. GPT-5.4's $r_{\mathrm{r}}$ is rendered NA because its reasoning channel is encrypted ciphertext.}
\label{tab:per-model-metrics}
\end{table}

\end{appendices}


\bibliography{sn-bibliography}

\end{document}